\documentclass[11pt]{article}

\usepackage[final]{acl}

\usepackage{times}
\usepackage{enumitem}
\usepackage{latexsym}
\usepackage[T1]{fontenc}
\usepackage[utf8]{inputenc}
\usepackage{microtype}
\usepackage{inconsolata}
\usepackage{graphicx}
\usepackage{booktabs}
\usepackage{array}
\usepackage{amsmath}
\usepackage{amssymb}
\usepackage{xspace}
\usepackage{url}
\usepackage{multirow}
\usepackage{algorithm}
\usepackage{algpseudocode}
\usepackage{xcolor}

\algrenewcommand\algorithmicrequire{\textbf{Input:}}
\algrenewcommand\algorithmicensure{\textbf{Output:}}

\newcommand{\method}{\textsc{INO}\xspace}
\newcommand{\rag}{RAG\xspace}

\title{Iterate Until Retrieved: Factual Nugget Optimization for Discoverable Continual Corrections in Agentic RAG}

\author{
 \textbf{Moshe Hazoom\textsuperscript{1}} \quad 
 \textbf{Gal Patel\textsuperscript{1}} \quad
 \textbf{Alon Talmor\textsuperscript{1}} \quad
 \textbf{Tom Hope\textsuperscript{1,2}}
 \\
 \textsuperscript{1}Mosaic AI \quad
 \textsuperscript{2}The Hebrew University of Jerusalem \\
 \{moshe, gal, alon, tom.h\}@getmosaic.ai
}

\begin{document}
\maketitle

\begin{abstract}
Agentic retrieval-augmented generation (RAG) systems in business-to-business settings may often receive free-form response feedback. Rather than generic feedback signals such as style, preference, or overall response quality, we focus on actionable factual corrections. We identify these instances and convert them into compact knowledge-base entries, which we call \emph{factual nuggets}. We introduce \emph{Iterative Nugget Optimization} (\method), an index-time optimization method that uses the production agentic RAG as a test harness: it creates an initial nugget, probes it with the triggering query and paraphrases, reflects over failed retrieval and answer traces, and revises the nugget until it is discoverable. We evaluate \method with two production B2B knowledge-assistance agents across multiple companies that use our system: a product support agent that answers questions over company-specific knowledge bases, and a support ticket agent that assists support engineers. \method consistently improves results over baselines in terms of discoverability and usage of factual corrections, in automated and human evaluations. A controlled experiment in production shows that the nuggets improve user satisfaction with responses.
\end{abstract}

\section{Introduction}

B2B (business-to-business) enterprise knowledge-assistance AI agents increasingly answer complex questions with retrieval-augmented generation (RAG), in which the agent's response is grounded by retrieved company knowledge or domain-specific evidence \cite{Yip2026,chang2026karlknowledgeagentsreinforcement}. For example, these agents support product specialists, customer-success teams, and support engineers by grounding answers in customer-specific documentation, prior tickets, product behavior notes, and internal operational knowledge, including product documentation, support knowledge bases, issue-tracking histories, and connected enterprise repositories \cite{packowski2024enterpriseRAG,xu2024ragkgcustomer,baqar2026rag4ticketsaipoweredticketresolution}. This assistance helps teams triage technical issues, resolve integration or configuration problems, and draft concrete next steps for complex cases.

\begin{figure}
    \includegraphics[width=\linewidth]{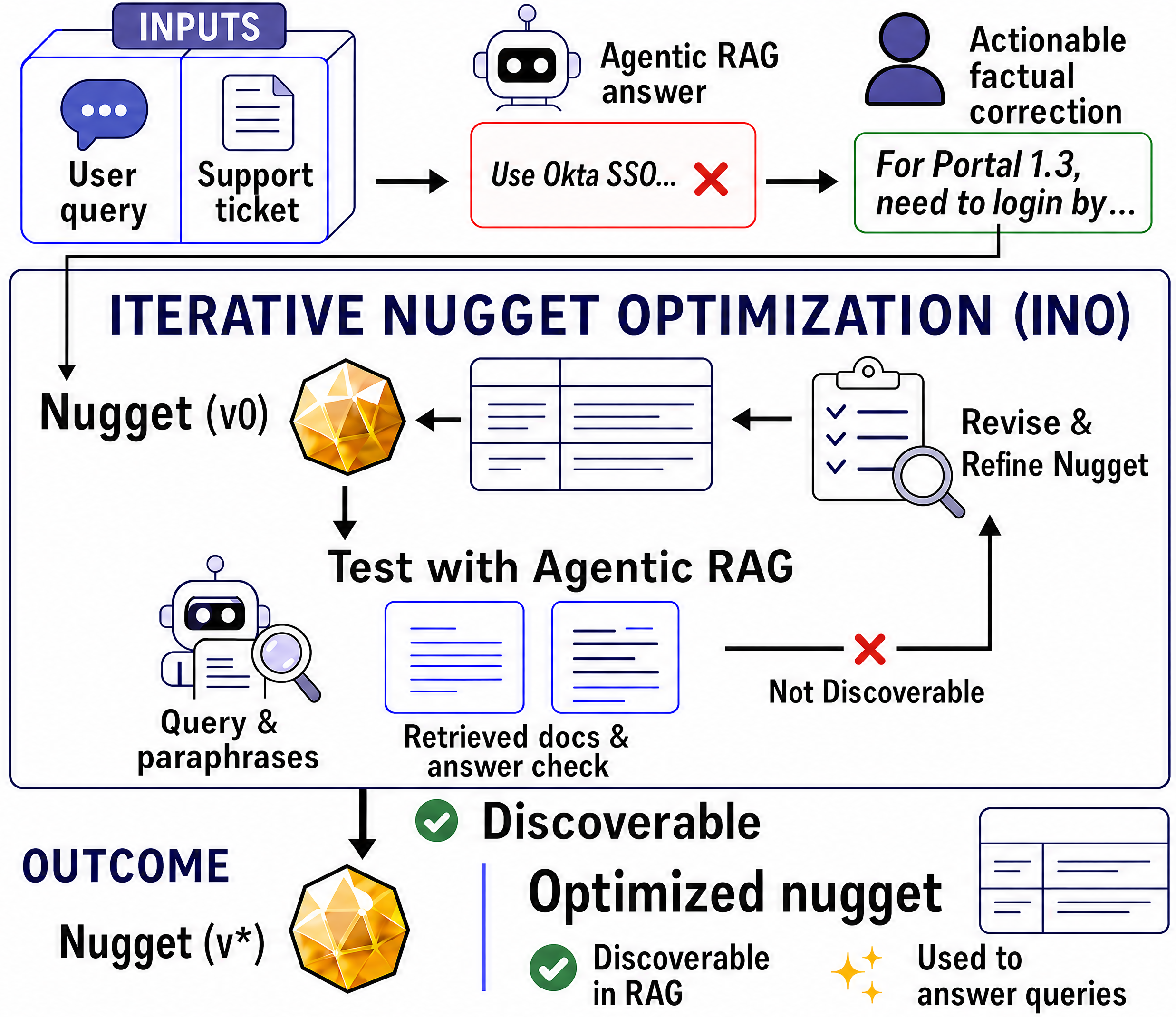}
    \caption{Overview of our method and setting. Factual corrections from users are detected and transformed into initial ``nuggets.'' We iteratively refine these nuggets using the agentic RAG system as a test harness, until they are discoverable and optimized for future queries.}
    \label{fig:teaser}
\end{figure}

Users of such systems often provide a continuous stream of feedback: thumbs-down, comments on missing information, and free-form corrections. This feedback is not homogeneous. Many negative signals may be vague (``not helpful''), subjective (tone or style), or local to a single conversation. A smaller but operationally important subset states a reusable factual correction: e.g., a product behavior, role restriction, naming clarification, or policy detail that should be available to future users.

This paper focuses on that subset. We first detect feedback that contains an actionable factual correction. Given this feedback, the goal is to produce a short, self-contained knowledge-base entry: a \emph{factual nugget} that future queries can discover and use. The problem is not simply to summarize the feedback. The nugget must encode the corrected fact conservatively (to avoid inserting wrong knowledge), include enough context to be reusable and standalone, and match future queries that may share little lexical overlap with the original query. It should also avoid becoming so broad that it is retrieved for unrelated traffic. A straightforward approach that simply inherits the wording and context of the original interaction is often too narrow. Future users ask from different roles, mention different issues, or use company-specific vocabulary.

\begin{figure*}[t]
  \centering
  \includegraphics[width=\linewidth]{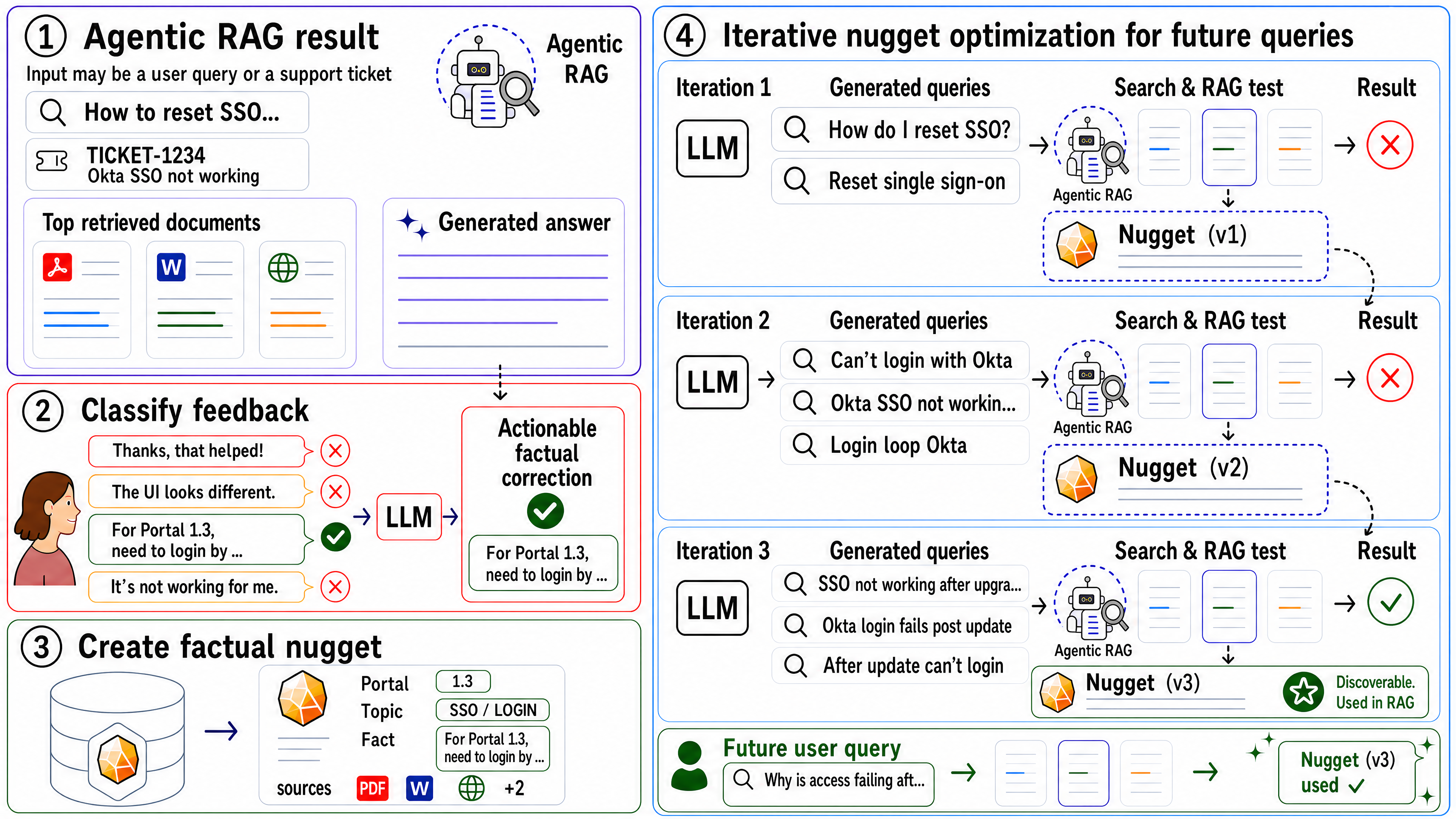}
  \caption{Overview of \method. User feedback is first filtered for actionable factual corrections. A correction is converted into a structured factual nugget. The nugget is then optimized with the production agentic \rag stack in the loop: generated test queries are replayed, retrieved competitors and generated answers are inspected, and the nugget is revised until it becomes discoverable and usable for relevant future queries.}
  \label{fig:overview}
\end{figure*}

We introduce \emph{Iterative Nugget Optimization} (\method), illustrated in Figure~\ref{fig:teaser}. \method treats nugget construction as a measurable index-time optimization problem. Starting from a factual correction, it creates a candidate nugget and a small set of likely user queries. It then inserts the candidate into the relevant corpus and runs the same agentic \rag system that will serve future users. If the nugget is not retrieved or used for the triggering query and probe paraphrased queries, an LLM reflector inspects the failed queries, the retrieved competing documents, and the generated answers, then revises the nugget representation. Unlike recent iterative prompt and context optimization methods \citep{agrawal2026gepa,zhang2026agentic}, \method does not optimize the system prompt/context.  Prompt optimizers amortize changes across many future  tasks by changing instructions or context; \method amortizes work across future queries about one factual correction by changing the retrievable artifact. This keeps the production agent configuration fixed and makes the optimization local, auditable, and reversible at the nugget level. \textbf{In summary, our main contributions:}
\begin{enumerate}[wide, labelwidth=!, labelindent=0pt]
    \item We formulate the task of converting feedback on the responses of B2B AI knowledge-assistance agents into discoverable factual nuggets, separating factual corrections from generic feedback.
    
    \item We propose \method (Iterative Nugget Optimization), an agentic-\rag-in-the-loop method that reflects over retrieval and answer traces to iteratively refine the indexable nugget.
    
    \item We evaluate several methods on real production data from two agents across multiple companies served by our system: one where inputs are free-form questions, and the other where inputs are support tickets. Across all experiments, \method leads to large gains in nugget discoverability and usage in generated answers. \method has been launched in production, and a controlled experiment shows that the nuggets improve satisfaction with response quality.

\end{enumerate}

\section{Related Work}

\textbf{Document expansion for retrieval.}
Document expansion augments retrievable units with synthetic queries or related text \citep{DBLP:journals/corr/abs-1904-08375,nogueira2019from}. \citet{weller2024when} show that expansion can fail or even hurt depending on method, retriever, and dataset. Our setting differs because the unit being inserted is a short feedback-derived nugget, generated online for a specific customer corpus, and optimized on a per-nugget basis by iteratively executing queries on the downstream agentic RAG and reflecting on the results.

\textbf{LLM-augmented agent memory.}
Agent memory methods 
rewrite past interactions into self-contained memories, often adding metadata such as keywords, tags, and contextual descriptions \citep{xu2025amem}. Our setting differs in both source and target: we start from free-form feedback that must be filtered for actionable factual corrections, and we store company-owned factual nuggets in an enterprise KB rather than private episodic memories. \method also adds optimization downstream where each nugget is tested and revised against the unchanged production agentic RAG stack until it is discoverable and usable for relevant future queries.

\textbf{Feedback-driven RAG updates.}
PatchRAG \citep{bang2026feedback} studies synthetic feedback on academic benchmarks, assuming feedback is a full corrected answer stored in a parallel feedback memory. It does not evaluate real company data with user feedback, focus on factual corrections, or optimize feedback representation.
STACKFEED \citep{kirtania2025stackfeed} edits existing KB documents using oracle or expert feedback on downstream failures, such as compiler errors or test failures, in software-engineering and factual-QA benchmarks. In contrast, we use free-form end-user feedback from deployed B2B knowledge-assistance agents, create new atomic factual nuggets rather than editing customer source knowledge, and optimize those nuggets for discoverability and use by an unchanged production agentic RAG stack. Variant D in section \ref{sec:methods} serves as the closest instantiation of their shared
memory-writing step under our deployment constraints.

\section{Task and Production Setting}

\subsection{Factual correction feedback}

The deployment studied here contains two production \rag agents used by B2B SaaS customers. One agent answers free-form support questions over each customer's KB. The second agent reads support tickets and suggests responses. See Figure \ref{fig:overview} (1). Each customer has an independent corpus and agent configuration. Customer corpora have a median of approximately 450k documents and range up to several million, so each newly inserted nugget must compete for retrieval against a large pool of nearby, customer-owned content. 

The feedback stream is noisy and covers many potential types of feedback aside from reusable \emph{factual} corrections. We therefore define an \emph{actionable factual correction} as feedback that provides enough concrete information to write a reusable factual statement. An LLM-based actionability extraction step filters the stream and outputs a nugget only when this criterion is met (Figure \ref{fig:overview} (2)). On a 5,000-event production slice, the actionable rate is 42.3\% for the chat agent and 21.5\% for the ticket-transcript agent. On a separate human-labeled calibration set, the classifier reaches 96.7\% agreement with human labels; details are in Appendix~\ref{app:actionability}.

\subsection{Factual nuggets}

A factual nugget is a short, standalone KB document that encodes one corrected fact plus the minimal context needed for reuse. It typically contains a title, a one-to-three sentence body, and optional retrieval anchors. For example, a correction about SSO reset behavior should not merely state that a previous answer was wrong; it should encode the reusable fact, such as which portal version and user role the reset procedure applies to (Figure \ref{fig:overview} (3)).

Importantly, we create new nuggets rather than editing customer documents. First, customer KB documents are the customer-owned source of truth, often covering broad workflows; a feedback event usually concerns one narrow detail inside a broader topic. Second, local in-place edits are hard to validate globally: changing one sentence can alter the coherence of a multi-step procedure. A compact auxiliary nugget keeps the correction atomic while leaving source documents unchanged.

The system separates nugget creation from publication. Feedback-derived nuggets remain unpublished until approved by a customer administrator or knowledge owner, although customers may configure automatic publication. Each nugget retains its provenance—including the feedback event, triggering query, corrected answer, timestamp, customer, and designated owner—to support efficient review and accountability.

Like other corpus documents, published nuggets may become outdated; managing staleness in evolving corpora is a general retrieval problem beyond the scope of this work. However, because nuggets are atomic and auxiliary, they can be unpublished or deleted without modifying source documents or disrupting multi-step procedures. Their scope further limits risk: for example, a nugget restricted to ``portal v8.2'' becomes irrelevant to out-of-scope queries when superseded (§ \ref{subsec:variant-examples}).

\subsection{Agentic retrieval stack}

All experiments use the production stack without modification. The retrieval agent includes a query generator and iterative retrieval based on intermediate results. Retrieval is hybrid, combining dense embeddings fine-tuned on proprietary in-domain data with sparse representations (e.g., BM25), followed by a fine-tuned cross-encoder re-ranker. Re-ranked scores are calibrated by a small MLP into $[0,1]$, and documents above a confidence threshold are passed to the generator. We use top-$k=60$ before re-ranking. Across variants, the only changed artifact is the content of the newly inserted nugget.

\subsection{Evaluation}
We evaluate nugget construction at three levels.
\begin{itemize}[leftmargin=*,noitemsep]
    \item \textbf{Retrieval:} The nugget passes the re-ranker and is available to the generator.
    \item \textbf{Citation:} The nugget is cited in the final answer's sources, indicating that it was used.
    \item \textbf{Answer feedback compliance:} an LLM judge checks whether the answer addresses the correction in the original feedback. The judge is calibrated against human annotators (Section~\ref{sec:answer-level}).
\end{itemize}

\section{Methods}\label{sec:methods}

Let $n$ be the initial factual nugget extracted from feedback on query $q_0$. Nugget construction produces an artifact $n^\star$ containing the nugget text and, for some variants, additional retrieval anchors. We compare five strategies, including our method \method (Figure \ref{fig:overview} (4)). See examples in Sec. \ref{subsec:variant-examples}.

\textbf{A. Standard nugget.}\label{var:standard}
The baseline indexes the nugget as-is: $n^\star = n$. This captures the common memory-generation pattern of turning a feedback event into a short standalone fact without retrieval enhancement \citep{chhikara2025mem0}, and is the closest instantiation under our
constraints of the write-the-correction-as-a-record step shared by
\citet{bang2026feedback} and \citet{kirtania2025stackfeed}. Note that our agentic RAG system includes query-time expansion as default, so this baseline naturally includes queries that are expanded by an LLM.

\textbf{B. Trigger-query anchor.}
Appends a paraphrase of the original user query $q_0$ to the nugget, produced by an LLM.

\textbf{C. Synthetic query anchors.}
An LLM generates five future queries that the nugget should answer, appended to the nugget; a sample was manually inspected to verify alignment with $q_0$.

\textbf{D. Trigger plus synthetic anchors.}
Adding a paraphrase of $q_0$ and four other queries (\textbf{B} + \textbf{C}).

\textbf{E. Iterative Nugget Optimization (\method).} Starts from Variant D but closes the loop with the full agentic \rag system. It indexes the candidate in a customer corpus and evaluates it against a \emph{probe set} of 4--6 queries: $q_0$ together with 3--5 held-out paraphrases generated fresh for this evaluation. These probe queries are deliberately disjoint from the usage examples bundled into the nugget (Variant D): the usage examples live \emph{inside} the indexed document and act as retrieval anchors, while the probe queries stay \emph{outside} and test whether retrieval generalizes to unseen but semantically related phrasings, i.e., the condition the nugget will face in deployment. For each probe query, we record whether the nugget was retrieved and cited. When a check fails, an LLM reflector receives the failed query, retrieved competitors, generated answer, and current nugget. It revises the indexable nugget representation: title, body wording, and retrieval anchors, while preserving the factual correction. The revised candidate is re-indexed and retested, for at most $T=3$ iterations.

\begin{algorithm}[H]
\small
\raggedright
\caption{Iterative Nugget Optimization}
\label{alg:reflect}
\begin{algorithmic}[1]
\Require nugget $n$, trigger query $q_0$, max iterations $T$
\Ensure indexable form $n^\star$
\State $n^\star \gets \textsc{InitialExpansion}(n, q_0)$ \Comment{Variant D}
\State $Q \gets \{q_0\} \cup \textsc{Paraphrase}(q_0)$ \Comment{held-out probes, fixed}
\For{$t = 1, \ldots, T$}
    \State $\textsc{Index}(n^\star)$
    \State $F \gets \{(q, \textsc{RAG}(q)) : q \in Q,\ \neg\,\textsc{Pass}(n^\star, q)\}$
    \If{$F = \emptyset$}
        \State \textbf{break} \Comment{converged; stays indexed}
    \EndIf
    \State $\textsc{Unindex}(n^\star)$
    \State $n^\star \gets \textsc{Rewrite}(n^\star, F, q_0)$
\EndFor
\State \Return $n^\star$
\end{algorithmic}
\end{algorithm}
Here $\textsc{Pass}(n^\star, q)$ holds when the nugget is retrieved and cited
for probe query $q$; $F$ collects the probes that fail, paired with their
retrieval and answer traces.

The reflector is instructed not to add new factual claims beyond the correction and provided context. Its role is to make the correction more discoverable: e.g., add missing terminology, disambiguate roles or products, and distinguish the nugget from documents that incorrectly outranked it.

\section{Experimental Setup}

\textbf{Chat agent} The main evaluation uses 100 factual nuggets sampled from real negative-feedback events that passed the actionability filter. We sample roughly uniformly across seven customers. The nuggets span products, user roles, and topics within each customer's corpus. For each construction strategy, we generate $n^\star$ for all 100 nuggets, index them into a frozen copy of the corresponding customer corpus, replay queries through the same agent version (prompt, hyper-parameters, tools and configuration), and then delete the inserted nuggets before the next variant. Each variant is run three times because LLM-generated anchors vary between runs.

We note that for \method, 56\% of nuggets pass after the first iteration, 32\% after the second, and 12\% require the third; no evaluated nugget exhausts the budget without passing the retrieval check.

\textbf{Held-out queries} We evaluate generalization on 410 held-out queries constructed by two pipelines. The first mines real production queries: for each customer, we build a per-customer FAISS index over five months of production traffic, retrieve nearest neighbors for each triggering query, and keep high-scoring candidates after cross-encoder re-ranking. This yields 110 real historical paraphrases. The second pipeline generates three paraphrases per triggering query with a different LLM from the one used for nugget construction, grounded in domain vocabulary retrieved from the customer corpus. This yields 300 synthetic paraphrases. See Appendix~\ref{app:heldout} for implementation details.

\textbf{Support ticket agent transfer} We extend our evaluation to an agent that consumes support ticket transcripts, retrieves relevant documents, and produces a summary plus suggested next steps. We collect 160 ticket-level samples from real negative-feedback events that passed the actionability filter and apply the same methods.

\textbf{Negative control} We index nuggets and test 1,400 random customer queries, counting how often nuggets are retrieved and inspecting them.

\section{Results}

\begin{table*}[t]
\centering
\small
\begin{tabular}{llccccc}
\toprule
& & \multicolumn{2}{c}{\textbf{Historical queries (n=110)}} & \multicolumn{2}{c}{\textbf{LLM paraphrases (n=300)}} & \\
\cmidrule(lr){3-4}\cmidrule(lr){5-6}
\textbf{Variant} & \textbf{Strategy} & \textbf{Retrieval} & \textbf{Citation} & \textbf{Retrieval} & \textbf{Citation} \\
\midrule
A & Standard nugget & 52.3 $\pm$ 0.5 & 41.3 $\pm$ 1.6 & 64.1 $\pm$ 1.2 & 57.3 $\pm$ 1.1 \\
B & Trigger-query anchor & 67.0 $\pm$ 2.0 & 50.7 $\pm$ 1.5 & 76.1 $\pm$ 0.6 & 68.5 $\pm$ 2.3  \\
C & Synthetic anchors & 67.8 $\pm$ 1.3 & 52.0 $\pm$ 1.4 & 86.4 $\pm$ 0.7 & 77.0 $\pm$ 0.9  \\
D & Trigger + synthetic & 72.9 $\pm$ 2.2 & 58.9 $\pm$ 2.2 & 89.4 $\pm$ 0.3 & 81.5 $\pm$ 0.9  \\
E & \method & \textbf{77.3 $\pm$ 1.0} & \textbf{68.1 $\pm$ 1.8} & \textbf{93.7 $\pm$ 0.8} & \textbf{86.1 $\pm$ 1.1}  \\
\bottomrule
\end{tabular}
\caption{Held-out results split by query source. \method leads to large gains in both retrieval and citation.}
\label{tab:split}
\end{table*}

Table~\ref{tab:split} reports performance on held-out queries, while Table~\ref{tab:insample} shows results on the original queries (in-sample). Every variant improves over the standard nugget. The ordering is monotonic: trigger anchors help, synthetic anchors help more, combining them helps further, and \method performs best. The ranking is preserved on both retrieval and citation. On real historical queries, \method improves retrieval by 25.0 points and
citation by 26.8 points over the standard nugget baseline, with comparable
margins (29.6 and 28.8) on the synthetic subset.

The same ordering holds on the 110 real production queries alone. Real queries are uniformly harder: performance is up to 25 points lower than on LLM-generated paraphrases depending on the metric and variant. This makes the historical subset the more conservative estimate of field performance. Taken together, we show \method is generalizing across queries.

\begin{table}[t]
\centering
\small
\begin{tabular}{lcc}
\toprule
\textbf{Variant} & \textbf{Retrieval} & \textbf{Citation} \\
\midrule
A. Standard nugget & 67.6 $\pm$ 1.2 & 60.2 $\pm$ 1.0 \\
B. Trigger-query anchor & 79.1 $\pm$ 1.9 & 71.8 $\pm$ 2.3 \\
C. Synthetic anchors & 88.0 $\pm$ 1.2 & 79.5 $\pm$ 1.1 \\
D. Trigger + synthetic & 90.8 $\pm$ 1.1 & 81.9 $\pm$ 0.6 \\
E. \method & \textbf{97.0 $\pm$ 1.0} & \textbf{89.1 $\pm$ 1.6} \\
\bottomrule
\end{tabular}
\caption{In-sample evaluation on 100 triggering queries. Mean $\pm$ standard deviation over three runs.}
\label{tab:insample}
\end{table}

\begin{figure*}[t]
  \centering
  \includegraphics[width=\textwidth]{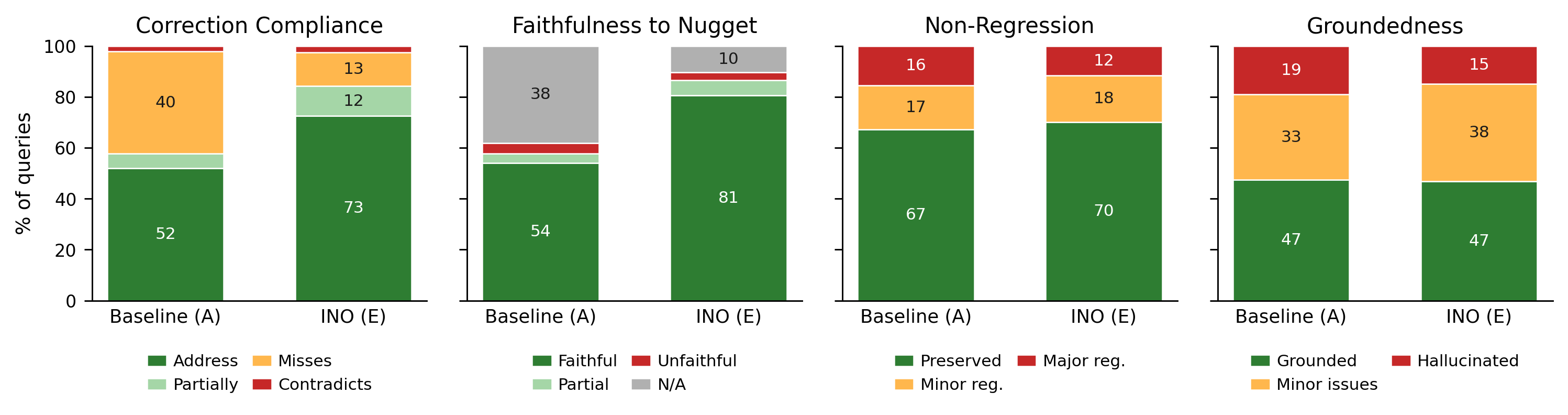}
  \caption{Answer-level judge results on held-out queries for the standard nugget baseline and \method. \method increases the share of answers that address the requested correction (73.4\% vs.\ 52.2\%; McNemar $p \approx 0.003$), and improves nugget availability without increasing major regressions or hallucinations. If the nugget is absent, the label is \emph{N/A}.}
  \label{fig:judge}
\end{figure*}

\subsection{Transfer to support tickets}

\begin{table}[h!]
\centering
\small
\begin{tabular}{lcc}
\toprule
\textbf{Variant} & \textbf{Retrieval} & \textbf{Citation} \\
\midrule
A. Standard nugget & 35.1 $\pm$ 0.9 & 29.7 $\pm$ 1.5 \\
B. Trigger-query anchor & 42.0 $\pm$ 1.9 & 37.8 $\pm$ 0.9 \\
C. Synthetic anchors & 48.8 $\pm$ 2.0 & 44.2 $\pm$ 1.3 \\
D. Trigger + synthetic & 63.3 $\pm$ 2.0 & 58.5 $\pm$ 2.5 \\
E. \method & \textbf{78.2 $\pm$ 1.8} & \textbf{70.4 $\pm$ 0.8} \\
\bottomrule
\end{tabular}
\caption{Support ticket agent evaluation on 160  samples. Mean $\pm$ standard deviation over three runs.}
\label{tab:tickets}
\end{table}

Table~\ref{tab:tickets} shows results on the support ticket agent. This setting is harder because the input is a multi-turn transcript rather than a clean question. The baseline retrieves the relevant nugget only 35.1\% of the time. The same ranking nevertheless holds, and \method improves retrieval by 43.1 points and citation by 40.7 points over the baseline.

The larger gains on tickets are consistent with the harder input distribution: transcripts contain filler, multiple issues, and vocabulary that differs from the original correction event. In this regime, per-nugget optimization has more room to help.

\subsection{Negative-control traffic}
Of 1,400 random queries, 20 (1.4\%) retrieved a newly inserted nugget and 12 (0.8\%) cited one. Manual inspection found no false positives. Each of the 20 retrievals was a semantically equivalent paraphrase of the original correction topic that happened to appear in the random sample. The implication is that the 1.4\% is not noise but generalization.

\subsection{Answer-level evaluation}
\label{sec:answer-level}

Retrieval and citation do not guarantee that the generated answer addresses the original correction. We therefore run an LLM-as-judge evaluation \citep{zheng2023judging} on held-out chat-agent answers for the baseline and \method. The judge sees the original question, original answer, user feedback, extracted nugget, retrieved context, and the answer. It labels four axes: correction compliance (does the answer materially reflect the correction), faithfulness to the nugget (does the answer adhere to the facts in the nugget), non-regression (is the new answer at least as good as the original without the nugget), and groundedness (does the answer make use of retrieved documents). See full metric definitions in Appendix \ref{app:judge}. The rubric is anchored to the user's correction because no gold answer exists for these production cases. We calibrate the judge against human annotators. A random sample of 100 judge decisions was independently labeled by three in-house annotators. Fleiss' $\kappa$ among annotators exceeds 0.75 on every axis. Using majority vote as the adjudicated label, Cohen's $\kappa$ between the judge and the adjudicated label exceeds 0.80 on every axis after a small prompt-calibration pass.

Figure~\ref{fig:judge} summarizes results. \method retrieves the nugget more often, and when it does, the generator uses it more faithfully. On correction compliance, \method addresses the correction in 73.4\% of held-out cases versus 52.2\% for the baseline, with the share of queries where the nugget was not retrieved ("misses") dropping from 40\% to 13\%. Conditional on the nugget being found, \method uses it faithfully in 81\% of cases versus 54.1\% for the baseline. Downstream, major regression errors decrease from 15.6\% to 11.6\%, and hallucinated groundedness labels decrease from 19.1\% to 14.8\%.

\subsection{Live Deployment Study}
\label{sec:post-deployment}
We deployed \method and hid the generated nuggets for a random 20\% of users over a 5-day window, comparing against the same window shifted seven days earlier. We restrict to queries with high propensity for nugget retrieval. We conduct a Difference-in-Differences (DiD) analysis on positive ratings (thumbs-up) given by users to responses. We  find that relative to the trend observed in the control group, hiding the nuggets caused an $11.44\%$ decrease in thumbs-up propensity for the blocked group (p-value=0.055).

\subsection{Example Nuggets Across the Five Variants}
\label{subsec:variant-examples}

To illustrate how the five construction strategies differ in practice, we show
one factual correction realized as $n^\star$ under each variant. The example is
sanitized and lightly paraphrased to avoid customer-identifying terminology;
proportions and structure match real production cases.

\paragraph{Source feedback event.}
Original user question $q_0$: \emph{``Why can't my analyst reset another
teammate's password from the admin panel?''} The agent answered that any user
with admin-panel access can reset passwords. The user marked the answer wrong
and commented that in portal v8.2 password reset is restricted to the
\emph{Workspace Owner} role; the Analyst role can view but not reset.

\paragraph{Extracted nugget $n$.}
\textbf{Title:} \emph{Password reset role restriction in portal v8.2.}
\textbf{Body:} In portal v8.2, only users with the Workspace Owner role can
reset another user's password from the admin panel. Users with the Analyst
role can view team members but cannot trigger a reset.

\paragraph{Variant A. Standard nugget.}
$n^\star = n$. No anchors, no expansion.

\paragraph{Variant B. Trigger-query anchor.}
$n^\star = n$ with one appended anchor: Analyst role cannot reset another user's password in the admin panel.

\paragraph{Variant C. Synthetic query anchors.}
$n^\star = n$ with five LLM-generated anchors covering plausible future
phrasings (no constraint to match $q_0$): (1) How do I reset a user's password in the admin panel; (2) Which role is required to reset passwords?; (3) Can analysts manage other users' credentials?; (4) Permissions for password reset in v8.2; (5) Workspace Owner vs Analyst capabilities.

\paragraph{Variant D. Trigger plus synthetic anchors.}
$n^\star = n$ with five anchors, the first constrained to be a paraphrase of
$q_0$: (1) Analyst role cannot reset another user's password in the admin panel; (2) Which role is required to reset passwords in v8.2?; (3) Can analysts manage other users' credentials?; (4) Permissions needed to trigger a password reset; (5) Workspace Owner vs Analyst capabilities.

\paragraph{Variant E. \method.}
Starting from Variant D, the agent fails the probe query
\emph{``user reset isn't working for my team lead''} because a generic
``Resetting your own password'' article outranks the nugget on terms like
\emph{reset} and \emph{isn't working}. The reflector inspects the failed query
and the competing document, then revises the indexable representation:

\noindent\textbf{Revised title:} \emph{Admin-panel password reset for other
users is restricted to Workspace Owner (portal v8.2).}

\noindent\textbf{Revised body:} In portal v8.2, resetting \emph{another user's}
password from the admin panel requires the Workspace Owner role. Users with
the Analyst role can view team members in the admin panel but the reset action
is hidden. This restriction does not apply to a user resetting their own
password from the login screen.

\noindent\textbf{Revised anchors:}

(1) Why can't an analyst reset a teammate's password from the admin panel? (2) Admin panel reset action greyed out for analyst role. (3) Which role can reset passwords for other users in v8.2? (4) Workspace Owner permissions for password reset. (5) Reset other user's password vs reset my own password.

The revision adds the disambiguating phrase \emph{another user's} (the wording
that distinguished the nugget from the self-reset competitor), surfaces the UI
symptom \emph{greyed out}, and explicitly contrasts the two reset paths. The
nugget passes all test queries on the next iteration.

\section{Conclusion}

We studied how to convert actionable factual correction feedback into factual nuggets that are discoverable and useful in production \rag systems. Our proposed Iterative Nugget Optimization (\method) loop uses the agentic \rag stack itself to test and refine each nugget before it is added to the retrievable corpus, while keeping the production agent configuration fixed. Across seven B2B customer deployments of two different agents, \method substantially improves discoverability and usage of factual corrections compared to baselines. The optimization is offline, with a per-correction cost of about \$0.31 and under two minutes per nugget in our setup, and adds no query latency after indexing, making it practical to run continuously. Our approach is currently running in production; a controlled experiment showed substantial impact on user satisfaction with response quality.

\section*{Limitations}

\paragraph{One retrieval architecture.}
All main experiments use our hybrid agentic RAG. Pure dense, pure sparse, or late-interaction retrievers may respond differently to retrieval anchors and iterative nugget refinement.

\paragraph{English-only evaluation.}
All evaluated corpora and queries are in English. The method itself is language-agnostic given multilingual retrieval and generation components, but the empirical claims here are English-only.

\paragraph{LLM dependence.}
Nugget extraction, anchor generation, reflection, and answer judging all use frontier LLMs. Different models may produce different anchors or judge labels. We did not run a full model ablation; we report the results of our actual deployed system.

\bibliography{custom}

\clearpage\appendix

\section{Prompt and Model Details}
\label{app:prompts}

The production prompts include customer-specific examples and are not reproduced verbatim. This appendix describes the prompt structure used in the experiments.

\paragraph{Actionability and nugget extraction.}
Input fields are conversation history (in case of a multi-turn), the final user question, the answer that received feedback, feedback type (either thumbs-up or thumbs-down), free-text feedback, and cited documents for the current turn. The model emits JSON with (i) a usefulness label, (ii) a boolean indicating whether the feedback can produce a factual nugget, (iii) a short reason, and (iv) when applicable, a title and body for the nugget. The prompt emphasizes that nuggets must contain reusable factual knowledge explicitly supported by the feedback, not a critique of the answer.

\paragraph{Synthetic anchors.}
For Variant C, the model receives the nugget and writes five concise queries that future users might ask whose answer is the nugget. The prompt asks for diverse surface forms and discourages copying nugget wording.

\paragraph{Trigger plus synthetic anchors.}
For Variant D, the model receives the nugget and triggering query $q_0$. It writes five anchors: the first is a non-verbatim paraphrase of $q_0$, and the other four cover alternative phrasings or nearby aspects of the same topic.

\paragraph{\method reflection.}
For \method, the reflector receives the current nugget, triggering query, probe queries, top retrieved documents for failed checks, whether the nugget was retrieved and cited, and the generated answer. It returns a revised indexable nugget representation. The prompt requires factual preservation: it may add terminology, clarify scope, or rewrite anchors, but it may not add new facts beyond the correction and supplied context.

\section{Held-out Set Construction}
\label{app:heldout}

For real-query mining, we collected five months of production queries per customer and embedded them with a Jina-v5-small dense encoder \citep{akram2026jina}. We built per-customer FAISS indexes and retrieved the top 50 nearest neighbors for each triggering query, keeping candidates with cosine similarity above 0.75. We then re-ranked with bge-reranker-v2-m3 \citep{li2023making,chen2024bge} and kept candidates above 0.85. This produced 110 historical paraphrases.

For synthetic paraphrases, we first ran an agentic \rag step against the customer corpus to collect domain vocabulary and relevant terminology. A second LLM, different from the one used for nugget construction, generated three realistic paraphrases per triggering query while using that vocabulary. We used OpenAI's GPT-5.4 with \textit{high} reasoning for this task. A random sample was manually spot-checked for topical consistency.

\section{Judge Rubric}
\label{app:judge}

The judge receives the original user question, the original answer, user feedback, the extracted nugget, retrieved context, and the answer. It emits JSON for four independent axes.

\paragraph{Correction compliance.}
Labels are: \emph{addresses} (the answer materially reflects the correction), \emph{partial} (mentions the corrective fact but incompletely), \emph{misses} (does not address the correction), and \emph{contradicts} (repeats or entails the corrected mistake). If the nugget is absent, the label is \emph{misses}.

\paragraph{Faithfulness to the nugget.}
If the nugget is in retrieved context, the judge labels the use as \emph{faithful}, \emph{partial}, or \emph{unfaithful}. If the nugget is absent, the label is \emph{N/A}.

\paragraph{Non-regression.}
Labels are \emph{preserved}, \emph{minor regression}, or \emph{major regression}, comparing correct content in the original answer with the new answer.

\paragraph{Groundedness.}
Labels are \emph{grounded}, \emph{minor issues}, or \emph{hallucinated}, based only on the retrieved context shown to the judge.

\section{Actionability Classifier}
\label{app:actionability}

The actionability filter is an LLM, specifically OpenAI's GPT-5.2 with \textit{medium} reasoning effort, used with a small set of in-context examples. It performs classification and extraction in one pass. The output JSON contains a \texttt{feedback\_usefulness} field, a \texttt{kb\_candidate} boolean, and an \texttt{article} object. We use a single pass to avoid disagreement between a classifier and a separate extractor and to reduce inference cost.

Human calibration used 200 production feedback events sampled across the seven customers. Two annotators labeled each event for whether it contained enough information to produce a reusable factual nugget; disagreements were adjudicated by a third annotator. The classifier agreed with the adjudicated label in 96.7\% of cases.

\section{Costs}

In our implementation, Variants A--D cost about \$0.012--\$0.0124 per nugget and complete in 12.9 seconds on average, dominated by extraction. \method adds roughly \$0.30 and 98.3 seconds per nugget, bringing total cost to about \$0.31 and latency to 111.2 seconds. This cost is offline: it occurs once per factual correction and adds no per-query latency after indexing. When this budget is acceptable, \method is the best-performing configuration across both agents. When cost is constrained, Variant D is a strong alternative, reaching 84.8\% / 75.1\% on held-out chat queries at a fraction of the cost.

\end{document}